\documentclass{ecai2014}
\usepackage{times}
\usepackage{graphicx}
\usepackage{latexsym}
\usepackage{hyperref}
                            
\usepackage{braket}

\usepackage[neverdecrease]{paralist}      

\usepackage{amsmath,amssymb}
\newtheorem{definition}{Definition}

\newtheorem{proposition}{Proposition}

\usepackage{enumitem}
\setlist{itemsep=0pt,topsep=3pt,parsep=0pt,partopsep=0pt}

\usepackage[disable]{todonotes}

\usepackage{ifthen}
\newcommand{\brifnotempty}[1]{\ifthenelse{\equal{#1}{}}{}{ \br{#1}}}
\newenvironment{lemma*}[2][]
	{\pagebreak[2] \par \noindent \textbf{Lemma~\ref{#2}\brifnotempty{#1}.}\it}{\par}
\newenvironment{theorem*}[2][]
	{\pagebreak[2] \par \noindent \textbf{Theorem~\ref{#2}\brifnotempty{#1}.}\it}{\par}
\newenvironment{proposition*}[2][]
	{\pagebreak[2] \par \noindent \textbf{Proposition~\ref{#2}\brifnotempty{#1}.}\it}{\par}
\newenvironment{corollary*}[2][]
	{\pagebreak[2] \par \noindent \textbf{Corollary~\ref{#2}\brifnotempty{#1}.}\it}{\par}

\newcommand{\tuple}[1] {\langle #1 \rangle}
\newcommand{\tpl}[1]{\br{#1}}

\newcommand{\im}{\leftarrow}

\newcommand{\obs}{\mathcal{O}}

\newcommand{\pws}[1]{2^{#1}}
\newcommand{\seq}[1]{\langle #1 \rangle}
\newcommand{\rlb}{\sigma}

\newcommand{\hrlb}{\hrl[\rlb]}
\newcommand{\hrl}[1][\rl]{H(#1)}
\newcommand{\lpif}{\leftarrow}
\newcommand{\brl}[1][\rl]{B(#1)}

\newcommand{\brlb}{\brl[\rlb]}
\newcommand{\brs}{\mathit{br}}
\newcommand{\mymodels}{\mathrel\mid\joinrel=}
\newcommand{\ent}{\mymodels}
\newcommand{\nf}{not\,}

\newcommand{\lgc}{L}

\newcommand{\kbs}{\mathbf{KB}}

\newcommand{\bss}{\mathbf{BS}}

\newcommand{\acc}{\mathbf{ACC}}

\newcommand{\bs}{S}

\newcommand{\brnot}{\mathop{\mathbf{not}\,}}
\newcommand{\sbrlits}{B}
\newcommand{\brlit}{L}
\newcommand{\mcs}{M}

\newcommand{\ctx}{C}
\newcommand{\kbmcs}{\mathit{kb}}

\newcommand{\fkb}[1]{F_{#1}}

\newcommand{\ofkb}[2]{F^{#1}_{#2}}

\newcommand{\indi}{i}

\newcommand{\emMCS}{\mbox{eMCS}}
\newcommand{\evolving}{evolving}

\newcommand{\nxt}{next}
\newcommand{\now}{now}

\newcommand{\OUT}{OUT}
\newcommand{\op}{Op}

\newenvironment{textitem}
	{\setlength{\pltopsep}{.7ex}\setlength{\plitemsep}{.7ex}\begin{compactitem}}
	{\end{compactitem}}

\newenvironment{textenum}[1][]
	{\setlength{\pltopsep}{.7ex}\setlength{\plitemsep}{.7ex}\begin{compactenum}[#1]}
	{\end{compactenum}}
	
\newcommand{\redl}{\mathit{red}}

\newcommand{\br}[1]{(#1)}

\newcommand{\geql}{\mathbf{GE}}

\newcommand{\wfs}{\mathbf{WFS}}

\begin{document}

\title{Towards Efficient Evolving Multi-Context Systems (Preliminary Report)}

\author{Ricardo Gon\c{c}alves \and Matthias Knorr \and Jo\~{a}o Leite
\institute{CENTRIA \& Departamento de Inform{\'a}tica, Faculdade Ci\^encias e Tecnologia,
Universidade Nova de Lisboa, email: rjrg@fct.unl.pt} }

\maketitle
\bibliographystyle{ecai2014}

\begin{abstract}
Managed Multi-Context Systems (mMCSs) provide a general framework for integrating knowledge represented in heterogeneous KR formalisms.  
Recently, \evolving\ Multi-Context Systems (\emMCS s) have been introduced as an extension of mMCSs that add the ability to both react to, and reason in the presence of commonly temporary dynamic observations, and evolve by incorporating new knowledge.
However, the general complexity of such an expressive formalism may simply be too high in cases where huge amounts of information have to be processed within a limited short amount of time, or even instantaneously.
In this paper, we investigate under which conditions \emMCS s may scale in such situations and we show that such polynomial \emMCS s can be applied in a practical use case.
\end{abstract}

\section{Introduction}


Multi-Context Systems (MCSs) were introduced in~\cite{BrewkaE07}, building on the work in~\cite{GiunchigliaS94,RoelofsenS05}, to address the need for a general framework that integrates  knowledge bases expressed in heterogeneous KR formalisms.
Intuitively, instead of designing a unifying language (see e.g., \cite{GoncalvesA10,MotikR10}, and \cite{KnorrAH11} with its reasoner NoHR \cite{IvanovKL13a}) to which other languages could be translated, in an MCS the different formalisms and knowledge bases are considered as modules, and means are provided to model the flow of information between them (cf.~\cite{AlbertiGGLS11,HomolaKLS12,HybridJLC13} and references therein for further motivation on hybrid languages and their connection to MCSs). 

More specifically, an MCS consists of a set of contexts, each of which is a knowledge base in some KR formalism, such that each context can access information from the other contexts using so-called bridge rules. 
Such non-monotonic bridge rules add its head to the context's knowledge base provided the queries (to other contexts) in the body are successful.
Managed Multi-Context Systems (mMCSs) were introduced in~\cite{BrewkaEFW11} to provide an extension of MCSs by allowing operations, other than simple addition, to be expressed in the heads of bridge rules. This allows mMCSs to properly deal with the problem of consistency management within contexts.

One recent challenge for KR languages is to shift from static application scenarios which assume a one-shot computation, usually triggered by a user query, to open and dynamic scenarios where there is a need to react and evolve in the presence of incoming information. Examples include EVOLP~\cite{AlferesBLP02}, Reactive ASP \cite{GebserGKS11,GebserGKOSS12}, C-SPARQL \cite{BarbieriBCVG10}, Ontology Streams \cite{LecueP13} and ETALIS \cite{AnicicRFS12}, to name only a few. 

Whereas mMCSs are quite general and flexible to address the problem of integration of different KR formalisms, they are essentially static in the sense that the contexts do not evolve to incorporate the changes in the dynamic scenarios. 
In such scenarios, new knowledge and information is dynamically produced, often from several different sources -- for example a stream of raw data produced by some sensors, new ontological axioms written by some user, newly found exceptions to some general rule, etc. 

To address this issue, two recent frameworks, \evolving\ Multi-Context Systems (\emMCS s)~\cite{GKL14} and reactive Multi-Context Systems (rMCSs)~\cite{Brewka13,Ellmauthaler13,BEP14} have been proposed sharing the broad motivation of designing general and flexible frameworks inheriting from \mbox{mMCSs} the ability to integrate and manage knowledge represented in heterogeneous KR formalisms, and at the same time be able to incorporate knowledge obtained from dynamic observations.

Whereas some differences set \emMCS s and rMCSs apart (see related work in Sec.~\ref{sec:conclusions}), the definition of \emMCS s is presented in a more general way.
That, however, means that, as shown in \cite{GKL14}, the worst-case complexity is in general high, which may be problematic in dynamic scenarios where the overall system needs to evolve and react interactively.  
This is all the more true for huge amounts of data -- for example raw sensor data is likely to be constantly produced in large quantities -- and systems that are capable of processing and reasoning with such data are required.

At the same time, \emMCS s inherit from MCSs the property that models, i.e., equilibria, may be non-minimal, which potentially admits that certain pieces of information are considered true based solely on self-justification.
As argued in \cite{BrewkaE07}, minimality may not always be desired, which can in principle be solved by indicating for each context whether it requires minimality or not.
Yet, avoiding self-justifications for those contexts where minimality is desired has not been considered in \emMCS s.

In this paper, we tackle these problems and, in particular, consider under which conditions reasoning with \evolving\ Multi-Context Systems can be done in polynomial time.
For that purpose, we base our work on a number of notions studied in the context of MCSs that solve these problems in this case \cite{BrewkaE07}.
Namely, we adapt the notions of minimal and grounded equilibria to \emMCS s, and subsequently a well-founded semantics, which indeed paves the way to the desired result.

The remainder of this paper is structured as follows. 
After introducing the main concepts regarding mMCSs in Sect.~\ref{sec:prelim}, in Sect.~\ref{sec:emMCS} we recall with more detail the framework of \emMCS s already introducing adjustments to achieve polynomial reasoning. 
Then, in Sect.~\ref{sec:useCase} we present an example use case, before we adapt and generalize notions from MCSs in Sect.~\ref{sec:GrEquilWFS} as outlined. 
We conclude in Sect.~\ref{sec:conclusions} with discussing related work and possible future directions.

\section{Preliminaries: Managed Multi-Context Systems}\label{sec:prelim}


Following \cite{BrewkaE07}, a multi-context system (MCS) consists of a collection of components, each of which contains knowledge represented in some \emph{logic}, defined as a triple $\lgc = \tuple{\kbs, \bss, \acc}$ where $\kbs$ is the set of well-formed knowledge bases of $\lgc$, $\bss$ is the set of possible belief sets, and $\acc: \kbs \rightarrow \pws{\bss}$ is a function describing the semantics of $\lgc$ by assigning to each knowledge base a set of acceptable belief sets. 
We assume that each element of $\kbs$ and $\bss$ is a set, and define $\fkb{}=\{s:s\in \kbmcs \wedge \kbmcs \in \kbs\}$. 

In addition to the knowledge base in each component, \emph{bridge rules} are used to interconnect the components, specifying what knowledge to assert in one component given certain beliefs held in the components of the MCS. 
Bridge rules in MCSs only allow adding information to the knowledge base of their corresponding context. 
In~\cite{BrewkaEFW11}, an extension, called managed Multi-Context Systems (mMCSs), is introduced in order to allow other types of operations to be performed on a knowledge base.
For that purpose, each context of an mMCS is associated with a \emph{management base}, which is a set of operations that can be applied to the possible knowledge bases of that context. Given a management base $OP$ and a logic $\lgc$, let $OF=\{op(s): op\in OP \wedge s\in \fkb{}\}$ be the \emph{set of operational formulas} that can be built from $OP$ and $\fkb{}$. 
Each context of an mMCS gives semantics to operations in its management base using a \emph{management function} over a logic $\lgc$ and a management base $OP$, $mng:\pws{OF}\times \kbs\rightarrow \kbs$, i.e., $mng(op,kb)$ is the knowledge base that results from applying the operations in $op$ to the knowledge base $kb$. 
Note that this is already a specific restriction in our case, as $mng$ commonly returns a (non-empty) set of possible knowledge bases for mMCS (and eMCS).
We also assume that $mng(\emptyset,kb)=kb$. 
Now, for a sequence of logics $\lgc = \seq{\lgc_1, \dotsc, \lgc_n}$ and a management base $OP_\indi$, an \emph{$\lgc_i$-bridge rule $\rlb$ over $\lgc$}, $1 \leq i \leq n$, is of the form $\hrlb \lpif \brlb$ where $\hrlb\in OF_\indi$ and $\brlb$ is a set of \emph{bridge literals} of the forms $(r:b)$ and $\brnot (r:b)$, $1 \leq r \leq n$, with $b$ a belief formula of $\lgc_r$.

A \emph{managed Multi-Context System} (mMCS) is a sequence $M=\tuple{\ctx_1,\ldots,\ctx_n}$, where each $\ctx_\indi$, $\indi\in\{1,\ldots,n\}$, called a \emph{managed context}, is defined as $\ctx_\indi=\tuple{\lgc_\indi,\kbmcs_\indi,\brs_\indi,OP_\indi, mng_\indi}$ where $L_\indi=\tuple{\kbs_{\indi},\bss_{\indi},\acc_{\indi}}$ is a logic, $kb_\indi\in \kbs_{\indi}$, $\brs_\indi$ is a set of $\lgc_\indi$-bridge rules, $OP_\indi$ is a management base, and $mng_\indi$ is a management function over $L_\indi$ and $OP_\indi$.
  Note that, for the sake of readability, we consider a slightly restricted version of mMCSs where $\acc_{\indi}$ is still a function and not a set of functions as for logic suites \cite{BrewkaEFW11}.

For an mMCS $\mcs = \seq{\ctx_1, \dotsc, \ctx_n}$, a \emph{belief state of $\mcs$} is a sequence $\bs = \seq{\bs_1, \dotsc, \bs_n}$ such that each $\bs_i$ is an element of $\bss_{\indi}$. For a bridge literal $(r:b)$, $\bs\ent (r:b)$ if $b \in \bs_r$ and $\bs \ent \brnot (r:b)$ if $b \notin \bs_r$; for a set of bridge literals $\sbrlits$, $\bs \ent \sbrlits$ if $\bs \ent\brlit$ for every $\brlit \in \sbrlits$.
We say that a bridge rule $\rlb$ of a context $\ctx_i$ is \emph{applicable given a belief state $\bs$ of $\mcs$} if $\bs$ satisfies $\brlb$. 
We can then define $app_\indi(\bs)$, the set of heads of bridge rules of $\ctx_i$ which are applicable in $\bs$, by setting $app_\indi(\bs)=\{\hrlb:\rlb\in \brs_\indi \wedge \bs\ent \brlb\}$.

Equilibria are belief states that simultaneously assign an acceptable belief set to each context in the mMCS such that the applicable operational formulas in bridge rule heads are taken into account.
Formally, a belief state $\bs = \seq{\bs_1, \dotsc, \bs_n}$ of an mMCS $\mcs$ is an \emph{equilibrium} of $\mcs$ if, for every $1 \leq i \leq n$, $\bs_i \in \acc_{\indi}(mng_\indi(app_\indi(\bs),\kbmcs_\indi))$.

\section{Evolving Multi-Context Systems}\label{sec:emMCS}


In this section, we recall \evolving\ Multi-Context Systems as introduced in \cite{GKL14} including some alterations that are in line with our intentions to achieve polynomial reasoning.
As indicated in \cite{GKL14}, we consider that some of the contexts in the MCS become so-called \emph{observation contexts} whose knowledge bases will be constantly changing over time according to the observations made, similar, e.g., to streams of data from sensors.\footnote{For simplicity of presentation, we consider discrete steps in time here.}

The changing observations then will also affect the other contexts by means of the bridge rules.
As we will see, such effect can either be instantaneous and temporary, i.e., limited to the current time instant, similar to (static) mMCSs, where the body of a bridge rule is evaluated in a state that already includes the effects of the operation in its head, or persistent, but only affecting the next time instant. 
To achieve the latter, we extend the operational language with a unary meta-operation $\nxt$ that can only be applied on top of operations.

\begin{definition}
Given a management base $OP$ and a logic $\lgc$, we define  $eOF$, the evolving operational language, 
as $eOF=OF\cup\{\nxt(op(s)):op(s)\in OF\}$.
 \end{definition}

We can now define \evolving\ Multi-Context Systems.

\begin{definition}
An \emph{\evolving\ Multi-Context System (\emMCS)} is a sequence $M_e=\tuple{\ctx_1,\ldots,\ctx_n}$, where each \emph{\evolving\ context} $\ctx_i$, $i\in\{1,\ldots,n\}$ is defined as $\ctx_i=\tuple{\lgc_\indi,\kbmcs_\indi,\brs_\indi,OP_\indi, mng_\indi}$ where 

\begin{itemize}
 \item $L_\indi=\tuple{\kbs_{\indi},\bss_{\indi},\acc_{\indi}}$ is a logic
 
 \item $kb_\indi\in \kbs_{\indi}$
 
 \item $\brs_\indi$ is a set of $\lgc_\indi$-bridge rules s.t.\ $\hrlb\in eOF_\indi$
 
 \item $OP_\indi$ is a management base
 
 \item $mng_\indi$ is a management function over $L_\indi$ and $OP_\indi$.
  
\end{itemize}
\end{definition}
As already outlined, \evolving\ contexts can be divided into regular \emph{reasoning contexts} and special \emph{observation contexts} that are meant to process a stream of observations which ultimately enables the entire \emMCS\ to react and evolve in the presence of incoming observations. 
To ease the reading and simplify notation, w.l.o.g., we assume that the first $\ell$ contexts, $0\leq\ell\leq n$, in the sequence $\tuple{\ctx_1,\ldots,\ctx_n}$ are observation contexts, and, whenever necessary, such an \emMCS\ can be explicitly represented by $\tuple{\ctx_1^o,\ldots,\ctx_\ell^o, \ctx_{\ell+1},\ldots,\ctx_n}$.

As for mMCSs, a \emph{belief state for} $M_e$  is a sequence $S=\tuple{S_{1},\ldots,S_{n}}$ such that, for each 
 $1\leq i\leq n$, we have $S_i\in \bss_{\indi}$. 

Recall that the heads of bridge rules in an \emMCS\ are more expressive than in an mMCS, since they may be of two types: those that contain $\nxt$ and those that do not. 
As already mentioned, the former are to be applied to the current knowledge base and not persist, whereas the latter are to be applied in the next time instant and persist.
Therefore, we distinguish these two subsets.

\begin{definition}
Let $M_e=\tuple{C_1,\ldots,C_n}$ be an \emMCS\ and $S$ a belief state for $M_e$.
Then, for each $1\leq i\leq n$, consider the following sets:
\begin{itemize}

 \item  $app_\indi^{\nxt}(S)=\{op(s):\nxt(op(s))\in app_\indi(S)\}$

\item  $app_\indi^{\now}(S)=\{op(s):op(s)\in app_\indi(S)\}$

\end{itemize}

\end{definition}

Note that if we want an effect to be instantaneous and persistent, then this can also be achieved using two bridge rules with identical body, one with and one without $\nxt$ in the head.

Similar to equilibria in mMCS, the (static) equilibrium is defined to incorporate instantaneous effects based on $app_\indi^{\now}(S)$ alone.

\begin{definition}\label{def:staticEquilibrium}
Let $M_e=\tuple{C_1,\ldots,C_n}$  be an \emMCS.
A belief state $S=\tuple{S_{1},\ldots,S_{n}}$  for $M_e$ is a static \emph{equilibrium of $M_e$} iff,  
for each $1\leq i\leq n$, we have $\bs_i\in\acc_\indi(mng_\indi(app^{\now}_\indi(S),kb_\indi))$.
\end{definition}
Note the minor change due to $mng$ now only returning one $\kbmcs$.

To be able to assign meaning to an \emMCS\ evolving over time, we introduce evolving belief states, which are sequences of belief states, each referring to a subsequent time instant.
\begin{definition}
Let $M_e=\tuple{C_1,\ldots,C_n}$ be an \emMCS.
 An \emph{evolving belief state} of size $s$ for $M_e$  is a sequence $S_e=\tuple{S^1,\ldots,S^s}$ where each $S^j$,  $1\leq j\leq s$,  is a belief state for $M_e$.
  \end{definition}

To enable an \emMCS\ to react to incoming observations and evolve, an observation sequence defined in the following has to be processed.
The idea is that the knowledge bases of the observation contexts $\ctx_i^o$ change according to that sequence.
\begin{definition}
Let $M_e=\tuple{\ctx_1^o,\ldots,\ctx_\ell^o, \ctx_{\ell+1},\ldots,\ctx_n}$ be an \emMCS.
An \emph{observation sequence} for $M_e$ is a sequence $Obs=\tuple{\obs^1,\ldots,\obs^m}$, such that, for each  $1\leq j \leq m$, $\obs^j=\tuple{o_1^j,\ldots, o_{\ell}^j}$ is an \emph{instant observation} with $o_{i}^j\in \kbs_\indi$ for each $1\leq i \leq \ell$. 
\end{definition}

To be able to update the knowledge bases in the \evolving\ contexts, we need one further notation.
Given an \evolving\ context $\ctx_\indi$ and $k\in \kbs_{\indi}$, we denote by $\ctx_\indi[k]$ the \evolving\ context in which $kb_\indi$ is replaced by $k$, i.e., $\ctx_\indi[k]=\tuple{L_\indi,k,\brs_\indi,OP_\indi,mng_\indi}$.

We can now define that certain evolving belief states are evolving equilibria of an \emMCS\ $M_e=\tuple{\ctx_1^o,\ldots,\ctx_\ell^o, \ctx_{\ell+1},\ldots,\ctx_n}$ given an observation sequence $Obs=\tuple{\obs^1,\ldots,\obs^m}$ for $M_e$. 
The intuitive idea is that, given an evolving belief state $S_e=\tuple{S^1,\ldots,S^s}$ for $M_e$, in order to check if $S_e$ is an evolving equilibrium, we need to consider a sequence of \emMCS s, $M^1,\ldots, M^s$ (each with $\ell$ observation contexts), representing a possible evolution of $M_e$ according to the observations in $Obs$, such that $S^j$ is a (static) equilibrium of $M^j$. 
The knowledge bases of the observation contexts  in $M^j$ are exactly their corresponding elements $o^j_i$ in $\obs^j$.
For each of the other contexts $C_i$, $\ell+1\leq i\leq n$, its knowledge base in $M^j$ is obtained from the one in $M^{j-1}$ by applying the operations in $app_\indi^{next}(S^{j-1})$. 

\begin{definition}\label{def:evolvingEquilibrium} 
Let $M_e=\tuple{\ctx_1^o,\ldots,\ctx_\ell^o, \ctx_{\ell+1},\ldots,\ctx_n}$ be an \emMCS, $S_e=\tuple{S^1,\ldots,S^s}$ an evolving belief state of size $s$ for $M_e$, and $Obs=\tuple{\obs^1,\ldots,\obs^m}$ an observation sequence for $M_e$ such that $m\geq s$. 
Then, $S_e$ is an \emph{evolving equilibrium} of size $s$ of $M_e$  given $Obs$ iff, for each $1\leq j \leq s$, $\bs^{j}$ is an equilibrium of 
$M^j=\tuple{C_1^o[o^j_1], \ldots, C_\ell^o[o^j_\ell],C_{\ell+1}[k_{\ell+1}^{j}],\ldots,C_n[k^j_n]}$
where, for each $\ell+1\leq i\leq n$, $k_{i}^{j}$ is defined inductively as follows:
\begin{itemize}
 \item $k_{i}^1=kb_\indi$
 
 \item $k_{i}^{j+1}=mng_\indi(app^{\nxt}_\indi(S^{j}),k^j_i)$ 
 \end{itemize}
\end{definition}
Note that $\nxt$ in bridge rule heads of observation contexts are thus without any effect, in other words, observation contexts can indeed be understood as managed contexts whose knowledge base changes with each time instant.

The essential difference to \cite{GKL14} is that the $k_i^{j+1}$ can be effectively computed (instead of picking one of several options), simply because $mng$ always returns one knowledge base.
The same applies in Def.~\ref{def:staticEquilibrium}.

As shown in \cite{GKL14}, two consequences of the previous definitions are that any subsequence of an evolving equilibrium is also an evolving equilibrium, and mMCSs are a particular case of \emMCS s.

\section{Use Case Scenario}\label{sec:useCase}


\newcommand{\prest}[1]{\ensuremath{\mathsf{#1}}}
\newcommand{\const}[1]{\ensuremath{\mathit{#1}}}
\newcommand{\lpnot}{\mathop{\sim\!}}
\newcommand{\vara}{\mathbf{x}}
\newcommand{\varb}{\mathbf{y}}
\newcommand{\varc}{\mathbf{z}}

\newcommand{\add}{add}
\newcommand{\rem}{rm}

In this section, we illustrate \emMCS s adapting a scenario on cargo shipment assessment taken from \cite{SlotaLS11}.

The customs service for any developed country assesses imported cargo for a variety of risk factors including terrorism, narcotics, food and consumer safety, pest infestation, tariff violations, and intellectual property rights.\footnote{The system described here is not intended to reflect the policies of any country or agency.}
Assessing this risk, even at a preliminary level, involves extensive knowledge about commodities, business entities, trade patterns, government policies and trade agreements. 
Some of this knowledge may be external to a given customs agency: for instance the broad classification of commodities according to the international Harmonized Tariff System (HTS), or international trade agreements. 
Other knowledge may be internal to a customs agency, such as lists of suspected violators or of importers who have a history of good compliance with regulations. 
While some of this knowledge is relatively stable, much of it changes rapidly. 
Changes are made not only at a specific level, such as knowledge about the expected arrival date of a shipment; but at a more general level as well. For instance, while the broad HTS code for tomatoes (0702) does not change, the full classification and tariffs for cherry tomatoes for import into the US changes seasonally.

Here, we consider an \emMCS\ $M_e=\tuple{\ctx_1^o,\ctx_2^o, \ctx_3,\ctx_4}$ composed of two observation contexts $\ctx_1^o$ and $\ctx_2^o$, and two reasoning contexts $\ctx_3$ and $\ctx_4$.
The first observation context is used to capture the data of passing shipments, i.e., the country of their origination, the commodity they contain, their importers and producers. 
Thus, the knowledge base and belief set language of $\ctx_1^o$ is composed of all the ground atoms over $\prest{ShpmtCommod}/2$, $\prest{ShpmtDeclHTSCode}/2$, $\prest{ShpmtImporter}/2$, $\prest{ShpmtCountry}/2$, $\prest{ShpmtProducer}/2$,  and also $\prest{GrapeTomato}/1$ and $\prest{CherryTomato}/1$. 
The second observation context $\ctx_2^o$ serves to insert administrative information and data from other institutions.
Its knowledge base and belief set language is composed of all the ground atoms over $\prest{NewEUMember}/1$, $\prest{Misfiling}/1$, and $\prest{RandomInspection}/1$.
Neither of the two observation contexts has any bridge rules.

The reasoning context $\ctx_3$ is an ontological Description Logic (DL) context that contains a geographic classification, along with information about producers who are located in various countries. It also contains a classification of commodities based on their harmonized tariff information (HTS chapters, headings and codes, cf.\ \url{http://www.usitc.gov/tata/hts}). 
We refer to \cite{EiterFSW10} and \cite{BrewkaEFW11} for the standard definition of $\lgc_3$; $\kbmcs_3$ is given as follows:

\begin{tabbing}
		fooooooooooooooooooooooo\=\kill
		$\prest{Commodity} \equiv (\exists \prest{HTSCode}.\top)$\\
		$\prest{EdibleVegetable}\equiv (\exists \prest{HTSChapter}.\set{\text{`07'}})$ \\
		$\prest{CherryTomato}\equiv (\exists \prest{HTSCode}.\set{\text{`07020020'}})$\\
		$\prest{Tomato}\equiv (\exists \prest{HTSHeading}.\set{\text{`0702'}})$ \\
		$\prest{GrapeTomato}\equiv (\exists \prest{HTSCode}.\set{\text{`07020010'}})$\\
		$\prest{CherryTomato} \sqsubseteq \prest{Tomato}$
		\> $\prest{CherryTomato} \sqcap \prest{GrapeTomato} \sqsubseteq \bot$\\
		$\prest{GrapeTomato} \sqsubseteq \prest{Tomato}$
		\> $\prest{Tomato} \sqsubseteq \prest{EdibleVegetable}$ \\
		$\prest{EURegisteredProducer}\equiv (\exists \prest{RegisteredProducer}.\prest{EUCountry})$ \\
		$\prest{LowRiskEUCommodity}\equiv (\exists \prest{ExpeditableImporter}.\top)\sqcap$ \\
		\>$(\exists \prest{CommodCountry}.\prest{EUCountry})$\\
		$\prest{EUCountry}(\const{portugal})$	
	    \> $\prest{RegisteredProducer}(\const{p_1}, \const{portugal})$\\
		$\prest{EUCountry}(\const{slovakia})$
		\> $\prest{RegisteredProducer}(\const{p_2}, \const{slovakia})$
\end{tabbing}

$OP_3$ contains a single $\add$ operation to add factual knowledge.
The bridge rules $\brs_3$ are given as follows:

\begin{tabbing}
fooooooo\=\kill
        $\add(\prest{CherryTomato}(\vara))\lpif (1\!:\!\prest{CherryTomato}(\vara))$\\
        $\add(\prest{GrapeTomato}(\vara))\lpif (1\!:\!\prest{GrapeTomato}(\vara))$\\
        $\nxt(\add(\prest{EUCountry}(\vara)))\lpif (2\!:\!\prest{NewEUMember}(\vara))$\\
        $\add(\prest{CommodCountry}(\vara, \varb))\lpif (1\!:\!\prest{ShpmtCommod}(\varc, \vara))$,\\
		\>	$(1\!:\!\prest{ShpmtCountry}(\varc, \varb))$ \\
		$\add(\prest{ExpeditableImporter}(\vara, \varb))\lpif (1\!:\!\prest{ShpmtCommod}(\varc, \vara))$,  \\
		\>	$(1\!:\!\prest{ShpmtImporter}(\varc, \varb)),(4\!:\!\prest{AdmissibleImporter}(\varb))$,\\
		\>	$(4\!:\!\prest{ApprovedImporterOf}(\varb, \vara))$
\end{tabbing} 
Note that $\kbmcs_3$ can indeed be expressed in the DL $\mathcal{EL}^{++}$ \cite{BBL05} for which standard reasoning tasks, such as subsumption, can be computed in PTIME.

Finally, $\ctx_4$ is a logic programming (LP) indicating information about importers, and about whether to inspect a shipment either to check for
compliance of tariff information or for food safety issues.
For $\lgc_4$ we consider that $\kbs_\indi$ the set of normal logic programs over a signature $\Sigma$, $\bss_\indi$ is the set of atoms over $\Sigma$, and $\acc_\indi(\kbmcs)$ returns returns a singleton set containing only the set of true atoms in the unique well-founded model.
The latter is a bit unconventional, since this way undefinedness under the well-founded semantics \cite{GelderRS91} is merged with false information.
However, as long as no loops over negation occur in the LP context (in combination with its bridge rules), undefinedness does not occur, and the obvious benefit of this choice is that computing the well-founded model is PTIME-data-complete \cite{DantsinEGV01}.
We consider $OP_4=OP_3$, and $\kbmcs_4$ and $\brs_4$ are given as follows:

\begin{tabbing}
		foooooooooooooooooooooooooooooooooooooooooooo\=\kill
		$\prest{AdmissibleImporter}(\vara)\lpif \lpnot \prest{SuspectedBadGuy}(\vara).$\\
		$\prest{PartialInspection}(\vara)\lpif \prest{RandomInspection}(\vara).$ \\
		$\prest{FullInspection}(\vara)\lpif \lpnot \prest{CompliantShpmt}(\vara).$\\
	    $\prest{SuspectedBadGuy}(\const{i_1}).$
\end{tabbing}

\begin{tabbing}
fooooooooo\=\kill
       $\nxt((\prest{SuspectedBadGuy}(\vara))\lpif (2\!:\!\prest{Misfiling}(\vara))$\\
       $\add(\prest{ApprovedImporterOf}(\const{i_2}, \vara))\lpif (3\!:\!\prest{EdibleVegetable}(\vara))$ \\
	   $\add(\prest{ApprovedImporterOf}(\const{i_3}, \vara))\lpif (1\!:\!\prest{GrapeTomato}(\vara))$ \\
       $\add(\prest{CompliantShpmt}(\vara))\lpif (1\!:\!\prest{ShpmtCommod}(\vara, \varb))$,\\
       \>$(3\!:\!\prest{HTSCode}(\varb, \varc)),(1\!:\!\prest{ShpmtDeclHTSCode}(\vara, \varc))$\\
	   $\add(\prest{RandomInspection}(\vara))\lpif (1\!:\!\prest{ShpmtCommod}(\vara, \varb))$,\\
	   \> $(2\!:\!\prest{Random}(\varb))$ \\
	   $\add(\prest{PartialInspection}(\vara))\lpif (1\!:\!\prest{ShpmtCommod}(\vara, \varb))$,\\
	   \>$\brnot(3\!:\!\prest{LowRiskEUCommodity}(\varb))$\\   
	   $\add(\prest{FullInspection}(\vara))\lpif (1\!:\!\prest{ShpmtCommod}(\vara, \varb))$,\\
	   \>$(3\!:\!\prest{Tomato}(\varb)), (1\!:\!\prest{ShpmtCountry}(\vara, \const{slovakia}))$
\end{tabbing} 	

Now consider the observation sequence $Obs = \tuple{\obs^1,\obs^2,\obs^3}$ where $o_1^1$ consists of the following atoms on $\const{s_1}$ (where $s$ in $s_1$ stands for shipment, $c$ for commodity, and $i$ for importer):
\begin{tabbing}
        fooooooooooooooooooooo\=\kill
        $\prest{ShpmtCommod}(\const{s_1}, \const{c_1})$
		\> $\prest{ShpmtDeclHTSCode}(\const{s_1}, \text{`07020010'})$ \\
		$\prest{ShpmtImporter}(\const{s_1}, \const{i_1})$
		\> $\prest{CherryTomato}(\const{c_1})$
\end{tabbing}
$o_1^2$ of the following atoms on $\const{s_2}$:	
\begin{tabbing}
        fooooooooooooooooooooo\=\kill	
		$\prest{ShpmtCommod}(\const{s_2}, \const{c_2})$
		\> $\prest{ShpmtDeclHTSCode}(\const{s_2}, \text{`07020020'})$ \\
		$\prest{ShpmtImporter}(\const{s_2}, \const{i_2})$
		\>$\prest{ShpmtCountry}(\const{s_2}, \const{portugal})$\\
		$\prest{CherryTomato}(\const{c_2})$
\end{tabbing}
and $o_1^3$ of the following atoms on $\const{s_3}$:
\begin{tabbing}
        fooooooooooooooooooooo\=\kill
		$\prest{ShpmtCommod}(\const{s_3}, \const{c_3})$
		\> $\prest{ShpmtDeclHTSCode}(\const{s_3}, \text{`07020010'})$ \\
		$\prest{ShpmtImporter}(\const{s_3}, \const{i_3})$
		\> $\prest{ShpmtCountry}(\const{s_3}, \const{portugal})$\\
		$\prest{GrapeTomato}(\const{c_3})$
		\> $\prest{ShpmtProducer}(\const{s_3}, \const{p_1})$
\end{tabbing}
while $o_2^1=o_2^3=\emptyset$ and $o_2^2=\{\prest{Misfiling}(\const{i_3})\}$.
Then, an evolving equilibrium of size 3 of $M_e$ given $Obs$ is the sequence $S_e=\tuple{S^1,S^2,S^3}$ such that, for each $1\leq j\leq 3$, $S^j=\tuple{S^j_1,S^j_2,S^j_3,S^j_4}$. 
 Since it is not feasible to present the entire $S_e$, we just highlight some interesting parts related to the evolution of the system.
E.g., we have that $\prest{FullInspection}(\const{s_1})\in S^1_4$ since the HTS code does not correspond to the cargo; no inspection on $\const{s_2}$ in $S^2_4$ since the shipment is compliant, $\const{c_2}$ is a EU commodity, and $\const{s_2}$ was not picked for random inspection; and $\prest{PartialInspection}(\const{s_3})\in S^3_4$, even though $\const{s_3}$ comes from a EU country, because $\const{i_3}$ has been identified at time instant $2$ for misfiling, which has become permanent info available at time $3$.
\section{Grounded Equilibria and Well-founded Semantics}\label{sec:GrEquilWFS}

Even if we only consider MCSs $M$, which are static and where an implicit $mng$ always returns precisely one knowledge base, such that reasoning in all contexts can be done in PTIME, then deciding whether $M$ has an equilibrium is in NP \cite{BrewkaE07, BrewkaEFW11}.
The same result necessarily also holds for \emMCS s, which can also be obtained from the considerations on \emMCS s \cite{GKL14}.

A number of special notions were studied in the context of MCSs that tackle this problem \cite{BrewkaE07}.
In fact, the notion of minimal equilibria was introduced with the aim of avoiding potential self-justifications.
Then, grounded equilibria as a special case for so-called reducible MCSs were presented for which the existence of minimal equilibria can be effectively checked.
Subsequently, a well-founded semantics for such reducible MCSs was defined under which an approximation of all grounded equilibria can be computed more efficiently.
In the following, we transfer these notions from static MCSs in \cite{BrewkaE07} to dynamic \emMCS s and discuss under which (non-trivial) conditions they can actually be applied.

Given an \emMCS\ $M_e=\tuple{\ctx_1,\ldots,\ctx_n}$, we say that a static equilibrium $\bs=\seq{\bs_1, \dotsc, \bs_n}$ is \emph{minimal} if there is no equilibrium $\bs' = \seq{\bs_1', \dotsc, \bs_n'}$ such that $\bs_i' \subseteq \bs_i$ for all $i$ with $1 \leq i \leq n$ and $\bs_j' \subsetneq \bs_j$ for some $j$ with $1 \leq j \leq n$.

This notion of minimality ensures the avoidance of self-justifications in evolving equilibria.
The problem with this notion in terms of computation is that such minimization in general adds an additional level in the polynomial hierarchy.
Therefore, we now formalize conditions under which minimal equilibria can be effectively checked.
The idea is that the grounded equilibrium will be assigned to an \emMCS\ $M_e$ if all the logics of all its contexts can be reduced to special monotonic ones using a so-called reduction function.
In the case where the logics of all contexts in $M_e$ turn out to be monotonic, the minimal equilibrium will be unique.

Formally, a logic $\lgc =\tpl{\kbs, \bss, \acc}$ is \emph{monotonic} if
\begin{textenum}[1.]
	\item $\acc(\kbmcs)$ is a singleton set for each $\kbmcs \in \kbs$, and

	\item $\bs \subseteq \bs'$ whenever $\kbmcs \subseteq \kbmcs'$,
		$\acc(\kbmcs) = \set{\bs}$, and $\acc(\kbmcs') = \set{\bs'}$.
\end{textenum}

Furthermore, $\lgc = \tpl{\kbs, \bss, \acc}$ is \emph{reducible} if for some
$\kbs^* \subseteq \kbs$ and some reduction function $\redl : \kbs \times \bss
\rightarrow \kbs^*$,
\begin{textenum}[1.]
	\item the restriction of $\lgc$ to $\kbs^*$ is monotonic,

	\item for each $\kbmcs \in \kbs$, and all $\bs, \bs' \in \bss$:
		\begin{textitem}
			\item $\redl(\kbmcs, \bs) = \kbmcs$ whenever $\kbmcs \in \kbs^*$,

			\item 
				$\redl(\kbmcs, \bs) \subseteq \redl(\kbmcs, \bs')$ whenever $\bs'
				\subseteq \bs$,

			\item $\bs \in \acc(\kbmcs)$ iff $\acc(\redl(\kbmcs, \bs)) = \set{\bs}$.
		\end{textitem}
\end{textenum}

Then, an evolving context $\ctx = \tpl{\lgc,\kbmcs,\brs,OP, mng}$ is \emph{reducible} if its logic $\lgc$ is reducible and, for all $op\in\ofkb{OP}{\lgc}$ and all belief sets $\bs$, $\redl(mng(op,\kbmcs), \bs) = mng(op,\redl(\kbmcs, \bs))$.

An \emMCS\ is \emph{reducible} if all of its contexts are. 
Note that a context is reducible whenever its logic $\lgc$ is monotonic. In this case $\kbs^*$ coincides with $\kbs$ and $\redl$ is the identity with respect to the first argument. 

As pointed out in \cite{BrewkaE07}, reducibility is inspired by the reduct in (non-monotonic) answer set programming.
The crucial and novel condition in our case is the one that essentially says that the reduction function $\redl$ and the management function $mng$ can be applied in an arbitrary order.
This may restrict to some extent the sets of operations $OP$ and $mng$, but in our use case scenario in Sect.~\ref{sec:useCase}, all contexts are indeed reducible.

A particular case of reducible \emMCS s, definite \emMCS s, does not require the reduction function and admits the polynomial computation of minimal evolving equilibria as we will see next.
Namely, a reducible \emMCS\ $\mcs_e = \seq{\ctx_1, \dotsc, \ctx_n}$ is
\emph{definite} if
\begin{textenum}[1.]
	\item none of the bridge rules in any context contains $\brnot$,
	\item for all $i$ and all $\bs \in \bss_i$, $\kbmcs_i = \redl_i(\kbmcs_i,\bs)$.
\end{textenum}
In a definite \emMCS, bridge rules are monotonic, and knowledge bases are already in reduced form. 
Inference is thus monotonic and a unique minimal equilibrium exists. 
We take this equilibrium to be the grounded equilibrium.
	Let $\mcs_e$ be a definite \emMCS. A belief state $\bs$ of $\mcs_e$ is the
	\emph{grounded equilibrium of $\mcs_e$}, denoted by $\geql(\mcs_e)$, if $\bs$ is
	the unique minimal (static) equilibrium of $\mcs_e$.
This notion gives rise to evolving grounded equilibria.
\begin{definition}
Let $M_e=\tuple{\ctx_1,\ldots,\ctx_n}$ be a definite \emMCS, $S_e=\tuple{S^1,\ldots,S^s}$ an evolving belief state of size $s$ for $M_e$, and $Obs=\tuple{\obs^1,\ldots,\obs^m}$ an observation sequence for $M_e$ such that $m\geq s$. 
Then, $S_e$ is the \emph{evolving grounded equilibrium} of size $s$ of $M_e$  given $Obs$ iff, for each $1\leq j \leq s$, $\bs^{j}$ is a grounded equilibrium of $M^j$ defined as in Definition~\ref{def:evolvingEquilibrium}.
\end{definition}

Grounded equilibria for definite \emMCS s can indeed be efficiently computed following \cite{BrewkaE07}.
The only additional requirement is that all operations $op\in OP$ are \emph{monotonic}, i.e., for $\kbmcs$, we have that $\kbmcs\subseteq mng(op(s),\kbmcs)$.
Note that this is indeed a further restriction and not covered by reducible \emMCS s.
Now, for $1\leq i\leq n$, let $\kbmcs_\indi^0=\kbmcs_\indi$ and define, for each successor ordinal $\alpha+1$,
\[
\kbmcs_\indi^{\alpha+1} = mng(app_\indi^{\now}(E^\alpha),\kbmcs_\indi^{\alpha}), 
\]
where $E^\alpha=(E^\alpha_1,\ldots,E^\alpha_n)$ and $\acc_\indi(\kbmcs_i^\alpha)=\{E^\alpha_\indi\}$.
Furthermore, for each limit ordinal $\alpha$, define $\kbmcs^\alpha_\indi=\bigcup_{\beta\leq\alpha}\kbmcs^\beta_\indi$, and let $\kbmcs^\infty_\indi=\bigcup_{\alpha>0}\kbmcs_\indi^\alpha$.
Then Proposition 1 \cite{BrewkaE07} can be adapted:

\begin{proposition}
Let $M_e=\tuple{\ctx_1,\ldots,\ctx_n}$ be a definite \emMCS\ s.t.\ all $OP_\indi$ are monotonic.
A belief state $S=\tuple{S_1,\ldots,S_n}$ is the grounded equilibrium of $M_e$ iff $\acc_\indi(\kbmcs^\infty_\indi)=\{S_\indi\}$, for $1\leq i\leq n$.
\end{proposition}
As pointed out in \cite{BrewkaE07}, for many logics, $\kbmcs^\infty_\indi=\kbmcs^\omega_\indi$ holds, i.e., the iteration stops after finitely many steps.
This is indeed the case for the use case scenario in Sect.~\ref{sec:useCase}.

For evolving belief states $S_e$ of size $s$ and an observation sequence $Obs$ for $M_e$, this proposition yields that the evolving grounded equilibrium for definite \emMCS s can be obtained by simply applying this iteration $s$ times.

Grounded equilibria for general \emMCS s are defined based on a reduct which
generalizes the Gelfond-Lifschitz reduct to the multi-context case:

\begin{definition}
	Let $\mcs_e = \seq{\ctx_1, \dotsc, \ctx_n}$ be a reducible \emMCS\ and $\bs =
	\seq{\bs_1, \dotsc, \bs_n}$ a belief state of $\mcs_e$. The \emph{$\bs$-reduct
	of $\mcs_e$} is defined as
	$
		\mcs_e^\bs = \seq{\ctx_1^\bs, \dotsc, \ctx_n^\bs}
	$
	where, for each $\ctx_i = \tuple{\lgc_\indi,\kbmcs_\indi,\brs_\indi,OP_\indi, mng_\indi}$, we define
	$\ctx_i^\bs = \tpl{\lgc_i, \redl_i(\kbmcs_i, \bs_i), \brs_i^\bs,OP_\indi, mng_\indi}$. Here,
	$\brs_i^\bs$ results from $\brs_i$ by deleting all
	\begin{textenum}[1.]
		\item rules with $\brnot (r:p)$ in the body such that $\bs \ent (r:p)$,
			and

		\item $\brnot$ literals from the bodies of remaining rules.
	\end{textenum}
\end{definition}

For each reducible \emMCS\ $\mcs_e$ and each belief set $\bs$, the $\bs$-reduct of $\mcs_e$ is
definite. We can thus check whether $\bs$ is a grounded equilibrium in the
usual manner:

\begin{definition}
	Let $\mcs_e$ be a reducible \emMCS\ such that all $OP_\indi$ are monotonic. A belief state $\bs$ of $\mcs_e$ is a
	\emph{grounded equilibrium of $\mcs_e$} if $\bs$ is the grounded equilibrium
	of $\mcs_e^\bs$, that is $\bs = \geql(\mcs_e^\bs)$.
\end{definition}

The following result generalizes Proposition 2 from \cite{BrewkaE07}.

\begin{proposition}
Every grounded equilibrium of a reducible \emMCS\ $M_e$ such that all $OP_\indi$ are monotonic is a minimal equilibrium of $M_e$.
\end{proposition}

This can again be generalized to evolving grounded equilibria.

\begin{definition}
Let $M_e=\tuple{\ctx_1,\ldots,\ctx_n}$ be a normal, reducible \emMCS\ such that all $OP_\indi$ are monotonic, $S_e=\tuple{S^1,\ldots,S^s}$ an evolving belief state of size $s$ for $M_e$, and $Obs=\tuple{\obs^1,\ldots,\obs^m}$ an observation sequence for $M_e$ such that $m\geq s$. 
Then, $S_e$ is the \emph{evolving grounded equilibrium} of size $s$ of $M_e$  given $Obs$ iff, for each $1\leq j \leq s$, $\bs^{j}$ is the grounded equilibrium of $(M^j)^{S^j}$ with $M^j$ defined as in Definition~\ref{def:evolvingEquilibrium}.
\end{definition}

This computation is still not polynomial, since, intuitively, we have to guess and check the (evolving) equilibrium, which is why the well-founded semantics for reducible \emMCS s $M_e$ is introduced following \cite{BrewkaE07}.
Its definition is based on the operator $\gamma_{M_e}(S)=\geql(M_e^S)$, provided $\bss_\indi$ for each logic $\lgc_\indi$ in all the contexts of $M_e$ has a least element $S^*$.
Such \emMCS s are called \emph{normal}.

The following result can be straightforwardly adopted from \cite{BrewkaE07}.

\begin{proposition}
Let $M_e=\tuple{\ctx_1,\ldots,\ctx_n}$ be a reducible \emMCS\ such that all $OP_\indi$ are monotonic. Then $\gamma_{M_e}$ is antimonotone.
\end{proposition}

As usual, applying $\gamma_{M_e}$ twice yields a monotonic operator.
Hence, by the Knaster-Tarski theorem, $(\gamma_{M_e})^2$ has a least fixpoint which determines the well-founded semantics.

\begin{definition}
Let $M_e=\tuple{\ctx_1,\ldots,\ctx_n}$ be a normal, reducible \emMCS\ such that all $OP_\indi$ are monotonic.
The well-founded semantics of $M_e$, denoted $\wfs(M)$, is the least fixpoint of $(\gamma_{M_e})^2$.
\end{definition} 

Starting with the least belief state $S^*=\tuple{S^*_1,\ldots,S^*_n}$, this fixpoint can be iterated, and the following correspondence between $\wfs(M_e)$ and the grounded equilibria of $M_e$ can be shown.

\begin{proposition}
Let $M_e=\tuple{\ctx_1,\ldots,\ctx_n}$ be a normal, reducible \emMCS\ such that all $OP_\indi$ are monotonic, $\wfs(M_e)=\tuple{W_1,\ldots W_n}$, and $S=\tuple{S_1,\ldots,S_n}$ a grounded equilibrium of $M_e$.
Then $W_\indi\subseteq S_\indi$ for $1\leq i\leq n$.
\end{proposition}

The well-founded semantics can thus be viewed as an approximation of the belief state representing what is accepted in all grounded equilibria, even though $\wfs(M_e)$ may itself not necessarily be an equilibrium.
Yet, if all $\acc_\indi$ deterministically return one element of $\bss_\indi$ and the \emMCS\ is acyclic (i.e., no cyclic dependencies over bridge rules exist between beliefs in the \emMCS\, see \cite{GKL14}), then the grounded equilibrium is unique and identical to the well-founded semantics.
This is indeed the case for the use case in Sect.~\ref{sec:useCase}. 

As before, the well-founded semantics can be generalized to evolving belief states.

\begin{definition}
Let $M_e=\tuple{\ctx_1,\ldots,\ctx_n}$ be a normal, reducible \emMCS\ such that all $OP_\indi$ are monotonic,  and $Obs=\tuple{\obs^1,\ldots,\obs^m}$ an observation sequence for $M_e$ such that $m\geq s$. 
The \emph{evolving well-founded semantics} of $M_e$, denoted $\wfs_e(M)$, is the evolving belief state $S_e=\tuple{S^1,\ldots,S^s}$ of size $s$ for $M_e$ such that $\bs^{j}$ is the well-founded semantics of $M^j$ defined as in Definition~\ref{def:evolvingEquilibrium}.
\end{definition}

Finally, as intended, we can show that computing the evolving well-founded semantics of $M_e$ can be done in polynomial time under the restrictions established so far.
For analyzing the complexity in each time instant, we can utilize \emph{output-projected} belief states \cite{EiterFSW10}.
The idea is to consider only those beliefs that appear in some bridge rule body.
Formally, given an \evolving\ context $\ctx_i$ within $M_e=\tuple{C_1,\ldots,C_n}$, we can define $\OUT_\indi$ to be the set of all beliefs of $\ctx_\indi$ occurring in the body of some bridge rule in $M_e$.
The \emph{output-projection} of a belief state $\bs = \seq{\bs_1, \dotsc, \bs_n}$ of $M_e$ is the belief state $\bs' = \seq{\bs_1', \dotsc, \bs_n'}$, $\bs'_\indi=\bs_\indi\cap \OUT_\indi$, for $1\leq i\leq n$.

Following \cite{EiterFSW10,BrewkaEFW11}, we can adapt the \emph{context complexity} of $\ctx_i$ from \cite{GKL14} as the complexity of the following problem:
\begin{description}
\item[(CC)] Decide, given $\op_\indi\subseteq OF_\indi$ and $\bs_\indi'\subseteq \OUT_\indi$, if exist $\kbmcs_\indi'= mng_\indi(\op_\indi,\kbmcs_\indi)$ and $\bs_\indi\in\acc_\indi(\kbmcs_\indi')$ s.t.\ $\bs_\indi'=\bs_\indi\cap OUT_\indi$.
\end{description}

Problem (CC) can intuitively be divided into two subproblems: (MC) compute some $\kbmcs_\indi'= mng_\indi(\op_\indi,\kbmcs_\indi)$ and (EC) decide whether $\bs_\indi\in \acc(\kbmcs_\indi')$ exists s.t. $\bs_i'=\bs_i\cap OUT_i$.
Here, (MC) is trivial for monotonic operations, so (EC) determines the complexity of (CC). 

\begin{theorem}
Let $M_e=\tuple{\ctx_1,\ldots,\ctx_n}$ be a normal, reducible \emMCS\ such that all $OP_\indi$ are monotonic, $Obs=\tuple{\obs^1,\ldots,\obs^m}$ an observation sequence for $M_e$, and (CC) is in PTIME for all $\ctx_i$.
Then, for $s\leq m$, computing $\wfs_e^s(M_e)$ is in PTIME.
\end{theorem}
This, together with the observation that $\wfs_e(M_e)$ coincides with the unique grounded equilibrium, allows us to verify that computing the results in our use case scenario can be done in polynomial time.

\section{Related and Future Work}\label{sec:conclusions}


In this paper we have studied how \emMCS s can be revised in such a way that polynomial reasoning is possible, and we have discussed an example use case to which this result applies.
We have also investigated the adaptation of notions concerning minimality of (evolving) equilibria, and we observe that the notion of reducible \emMCS s is considerably restricted, but not to the same extent as the efficient computation of the well-founded semantics requires.
An open question is whether a more refined computation eventually tailored to less restrictive operations than considered here can be used to achieve similar results.

As mentioned in the Introduction, \emMCS s share the main ideas of reactive Multi-Context Systems sketched in \cite{Brewka13,Ellmauthaler13,BEP14} inasmuch as both aim at extending mMCSs to cope with dynamic observations. Three main differences distinguish them. First, whereas \emMCS s rely on a sequence of observations, each independent from the previous ones, rMCSs encode such sequences within the same observation contexts, with its elements being explicitly timestamped. This means that with rMCSs it is perhaps easier to write bridge rules that refer, e.g., to specific sequences of observations, which in \emMCS s would require explicit timestamps and storing the observations in some context, although at the cost that rMCSs need to deal with explicit time which adds an additional overhead. Second, since in rMCSs the contexts resulting from the application of the management operations are the ones that are used in the subsequent state, difficulties may arise in separating non-persistent and persistent effects, for example, allowing an observation to override some fact in some context while the observation holds, but without changing the context itself -- such separation is easily encodable in \emMCS s given the two kinds of bridge rules, i.e., with or without operator $next$. Finally, bridge rules with $\nxt$ allow for the specification of transitions based on the current state, such as the one encoded by the rule $\nxt(add(p))\im \nf p$, which do not seem possible in rMCSs. Overall, these differences indicate that an interesting future direction would be to merge both approaches, exploring a combination of explicitly timestamped observations with the expressiveness provided by operator $\nxt$.

Another framework that aims at modeling the dynamics of knowledge is that of evolving logic programs EVOLP~\cite{AlferesBLP02} focusing on updates of generalized logic programs. 
It is possible to show that EVOLP can be seen as a particular case of \emMCS s, using the operator $\nxt$ to capture the operator $assert$ of EVOLP.
We leave the details for an extended version. 
Closely related to EVOLP, hence to \emMCS, are the two frameworks of reactive ASP, one implemented as a solver \emph{clingo}~\cite{GebserGKS11} and one described in~\cite{Brewka13}. The system \emph{oclingo} extends an ASP solver for handling external modules provided at runtime by a controller. The output of these external modules can be seen as the observations of EVOLP. Unlike the observations in EVOLP, which can be rules, external modules in \emph{oclingo} are restricted to produce atoms so the evolving capabilities are very restricted. On the other hand, \emph{clingo} permits committing to a specific answer-set at each state, a feature that is not part of EVOLP, nor of \emMCS. Reactive ASP as described in \cite{Brewka13} can be seen as a more straightforward generalization of EVOLP where operations other than $assert$ for self-updating a program are permitted. Given the above mentioned embedding of EVOLP in \emMCS, and the fact that \emMCS s permit several (evolution) operations in the head of bridge rules, it is also not difficult to show that Reactive ASP as described in \cite{Brewka13} can be captured by \emMCS s.

Also, as already outlined in \cite{theOtherReactKnowPaper}, an important non-trivial topic is the study of the notion of minimal change within an evolving equilibrium. Whereas minimal change may be desirable to obtain more coherent evolving equilibria, there are also arguments against adopting a one-size-fits-all approach embedded in the semantics. 
Different contexts, i.e., KR formalisms, may require different notions of minimal change, or even require to avoid it -- e.g., suppose we want to represent some variable that can non-deterministically takes one of two values at each time instant:  minimal change could force a constant value.

Another important issue open for future work is a more fine-grained characterization of updating bridge rules (and knowledge bases) as studied in \cite{GKL-Clima14} in light of the encountered difficulties when updating rules \cite{SlotaL10,SlotaL12a,SlotaL14} and the combination of updates over various formalisms \cite{SlotaL12a,SlotaL12b}.

Also interesting is to study how to perform AGM style belief revision at the (semantic) level of the equilibria, as in Wang et al~\cite{WangZW13}, though different since knowledge is not incorporated in the contexts.

\ack We would like to thank the referees for their comments, which helped improve this paper considerably.
Matthias Knorr and Jo{\~a}o Leite were partially supported by FCT under project ``ERRO -- Efficient Reasoning with Rules and Ontologies'' ({PTDC}/{EIA}-{CCO}/{121823}/{2010}).
Ricardo Gon\c{c}alves was supported by FCT grant SFRH/BPD/47245/2008 and Matthias Knorr was also partially supported by FCT grant SFRH/BPD/86970/2012.

\end{document}